\documentclass[11pt, a4paper]{article}
\usepackage{graphicx}
\usepackage[english]{babel}
\usepackage{amsmath}
\usepackage{amsfonts}
\usepackage{amsthm}
\usepackage{appendix}
\usepackage{subcaption}
\usepackage{caption}
\usepackage{booktabs}
\usepackage{enumerate}
\usepackage{float}
\usepackage{multirow}
\usepackage{authblk} 
\usepackage[hidelinks]{hyperref} 
\usepackage[authoryear, round]{natbib}
\usepackage{mymacros}
\title{
	\vspace{-0.25cm} 
	\rule{\textwidth}{0.6mm} 
	\\[0.4cm]
	\textbf{Self-attentive Transformer for Fast and Accurate Postprocessing of Temperature and Wind Speed Forecasts\footnote{This is the version of the manuscript accepted for publication in \textit{Artificial Intelligence for the Earth Systems (AIES)}, American Meteorological Society (AMS). The final published version will be available at 10.1175/AIES-D-24-0127.1.}}\\[0.4cm]
	\rule{\textwidth}{0.6mm} 
	\vspace{-0.25cm} 
}

\author[1]{Aaron Van Poecke}
\author[2]{Tobias Sebastian Finn}
\author[3,4]{Ruoke Meng}
\author[1,3]{Joris Van den Bergh}
\author[3]{Geert Smet}
\author[3]{Jonathan Demaeyer}
\author[3,4]{Piet Termonia}
\author[1,3,5]{Hossein Tabari}
\author[1]{Peter Hellinckx}

\affil[1]{M4S, Faculty of Applied Engineering, University of Antwerp, Belgium}
\affil[2]{CEREA, École des Ponts and EDF R\&D, Île-de-France, France}
\affil[3]{Royal Meteorological Institute of Belgium, Brussels, Belgium}
\affil[4]{Department of Physics and Astronomy, Ghent University, Ghent, Belgium}
\affil[5]{United Nations University Institute for Water, Environment and Health, Hamilton, ON, Canada}

\affil[ ]{\vspace{1em}} 

\affil[ ]{\textit{Corresponding Author:} Aaron Van Poecke (\texttt{aaron.vanpoecke@uantwerpen.be})}

\date{} 
\begin{document}
	\maketitle
\begin{abstract}
Current postprocessing techniques often require separate models for each lead time and disregard possible inter-ensemble relationships by either correcting each member separately or by employing distributional approaches. In this work, we tackle these shortcomings with an innovative, fast and accurate Transformer which postprocesses each ensemble member individually while allowing information exchange across variables, spatial dimensions and lead times by means of multi-headed self-attention. Weather forecasts are postprocessed over 20 lead times simultaneously while including up to fifteen meteorological predictors. We use the EUPPBench dataset for training which contains ensemble predictions from the European Center for Medium-range Weather Forecasts' integrated forecasting system alongside corresponding observations. The work presented here is the first to postprocess the ten and one hundred-meter wind speed forecasts within this benchmark dataset, while also correcting two-meter temperature. Our approach significantly improves the original forecasts, as measured by the CRPS, with 16.5\% for two-meter temperature, 10\% for ten-meter wind speed and 9\% for one hundred-meter wind speed, outperforming a classical member-by-member approach employed as a competitive benchmark. Furthermore, being up to six times faster, it fulfills the demand for rapid operational weather forecasts in various downstream applications, including renewable energy forecasting.
\vspace{5em} 
\end{abstract}

\maketitle

\section{Introduction}
\subsection{Relevance and background}
Accurate weather forecasts are vital for society as a whole and indispensable for a myriad of segments of our economy, such as the agricultural, renewable energy and public health sectors. Inaccurate weather predictions can result in significant financial losses due to crop failure, severe negative health effects from poorly forecasted extreme events and incorrect predictions of renewable energy sources \citep{Challinor, Lazo, Mohanty,VanPoecke}. Weather forecasting remains, however, an exceptionally challenging task due to the chaotic nature of the atmosphere \citep{Patil}. Due to this complex nature and the societal significance of accurate weather forecasts, the last decades have seen continuous efforts into improving weather forecast accuracy \citep{Schultz}. Although Machine Learning (ML) models have recently achieved impressing forecasting accuracy \citep{Lam, Pathak, Bipangu}, operational weather forecasting still relies on Numerical Weather Prediction (NWP) models \citep{Rabier}. Inaccurate initial conditions and imperfect parameterizations of physical processess in NWP models lead to errors which accumulate over time, limiting the accuracy of these forecasts \citep{Bouallegue}. These limitations can partly be countered by a Monte-Carlo approach to cover the uncertainty in weather predictions, which results in not one, but an ensemble of deterministic predictions generated by perturbing initial conditions or model parameters \citep{Lewis}. These ensemble forecasts, despite introducing probability into weather forecasting, still suffer from inappropriate dispersion or systemic biases \citep{Vannitsem}. To address these shortcomings, statistical postprocessing techniques are employed, which essentially learn from discrepancies between historical forecasts and observations in order to correct future weather forecasts. Postprocessing nowadays forms an essential part of the forecasting chain operated by meteorological services on both a national and an international level \citep{Demaeyer}.\\~\\
Employing postprocessing methods in order to correct errors in forecasts has been an ongoing effort in the weather community for more than half a century \citep{Belgianmet}. Postprocessing techniques can be classified based on various characteristics, including the method used (e.g., statistical or machine learning), the type of forecast (deterministic or probabilistic), the assumption about the variable's distribution (parametric or nonparametric), and the number of variables involved (univariate or multivariate), among other factors \citep{Yang}. Another distinction can be made between methods that provide a predictive distribution as output (e.g. a normal distribution for temperature) and methods that correct each ensemble member separately, resulting in a corrected ensemble of the original size, known as member-by-member approaches \citep{VanSchaey}. In this work, we focus mainly on deep learning algorithms for the postprocessing of ensemble weather forecasts, which are inherently probabilistic. This is justified by the fact that ensemble forecasting has become the undeniable backbone of operational weather forecasting \citep{Lewis} and by the recent shift in the literature from classical, statistical techiques towards deep learning algorithms \citep{Hohlein}.
\subsection{Related work} \label{subsec:similar}
Earlier studies on postprocessing employing machine learning-based approaches have mainly utilized Neural networks (NNs) to postprocess ensemble weather forecasts of temperature and wind speed at station level \citep{Rasp, Bremnes, Schulz2}. More recently, there has been a surge in the use of more complex deep learning algorithms, such as Convolutional Neural Networks (CNNs) and Transformers employing the attention mechanism \citep{Veldkamp, Finn}. The ideas of the ensemble Kalman filter \citep{Evensen} and self-attention \citep{Vaswani} were bridged by \cite{Finn} to result in a self-attentive ensemble Transformer which allowed to capture the interaction of ensemble members of multiple predictors. A first test with this architecture was performed by \cite{Finn} to postprocess two-meter temperature for each lead time separately on a global scale with two-meter temperature, geopotential height and temperature at 500 hPa and 850 hPa respectively as predictors. This work was extended by \cite{Ashkboos}, who postprocessed multiple variables of the ENS-10 dataset, using the same Transformer architecture and compared its performance to other networks. For temperature at two meter and 850hPa the model of \cite{Finn} outperformed all others, while for geopotential height a LeNet-style network using CNNs achieved the best results \citep{Li}. Similarly, \cite{Bouallegue} extended the work of \cite{Finn} by combining his attention-based Transformer with the U-Net architecture for bias correction by \cite{Gronquist}, an architecture they named PoET, and compared their results to other architectures available in the literature. Their conclusions were similar to those of \cite{Ashkboos}, i.e. the Transformer outperformed all other methods for temperature variables, whereas the LeNet architecture performed slightly better for the geopotential height. Lastly, recent studies have also explored postprocessing across all lead times simultaneously. \cite{Wessel} investigated the lead-time dependence of statistical postprocessing methods for station-based temperature and ten-meter wind speed forecasts and found improvements in computation time without loss of performance when comparing to models trained separately per lead time. Finally, \cite{Mlakar} incorporated normalizing spine flows in their neural network and postprocessed temperature forecasts at station level for all lead times jointly, reporting that their method outperforms per-lead-time-based approaches.    

\subsection{Contribution of this work}  
The method proposed in this work is a modern adaptation of the self-attentive ensemble Transformer developed by \cite{Finn}, which has emerged among the best performing postprocessing method in several applications as described above. The application in this paper differs from previous, similar work employing Transformers in a number of ways. First of all, the architecture is adapted to postprocess weather forecasts for multiple lead times simultaneously, for 20 steps and up to five days in the future, instead of requiring a new model with separate estimated weights for every lead time, as is for example the case for the PoET architecture described above. This adaptation not only results in a fast-performing, state-of-the art deep learning-based postprocessing method but also allows different lead times to influence each other in the attention module described in section \ref{sec:methods}. Additionally, this results in gains in training time given that the training time per lead time decreases when postprocessing all lead times simultaneously, as detailed in section \ref{sec:methods}. While recent studies have also explored postprocessing across all lead times, our work differs by focusing on gridded forecasts rather than station-level data, and by including wind speed at both ten and one hundred meter. Furthermore, this architecture allows for the inclusion of multiple predictors at a relatively low computational cost. We regress up to fifteen predictors against the postprocessed variable, enabling information change across different variables. This work is the first to postprocess gridded wind speed within the EUPPBench dataset, at both ten and hundred meter, the latter of which has only been added in recent weeks, thereby facilitating future comparisons with other postprocessing methods. While temperature has been subject of many recent, Transformer-based postprocessing studies \citep{Finn, Ashkboos, Bouallegue}, wind speed, both at ten and hundred-meter, has been given less attention, despite its relevance for several applications, such as renewable energy, agriculture and safety \citep{Burlando, Zhang, Pinson}. To the best of our knowledge, this work is the first to apply a Transformer-based architecture for postprocessing wind speed at one hundred meter. The remainder of this paper is structured as follows: Section \ref{sec:methods} provides a detailed description of the Transformer, while also presenting the benchmark method and the dataset. Section \ref{sec:results} discusses the application of this method for the case studies of temperature and wind speed. Lastly, Section \ref{sec:conclusions} presents the discussion and future prospects.   
   
 \section{Methods and data} \label{sec:methods}
\subsection{Transformer} \label{subsec:Transformer}
\begin{figure*}[ht]
	\centering
	\includegraphics*[scale=0.65]{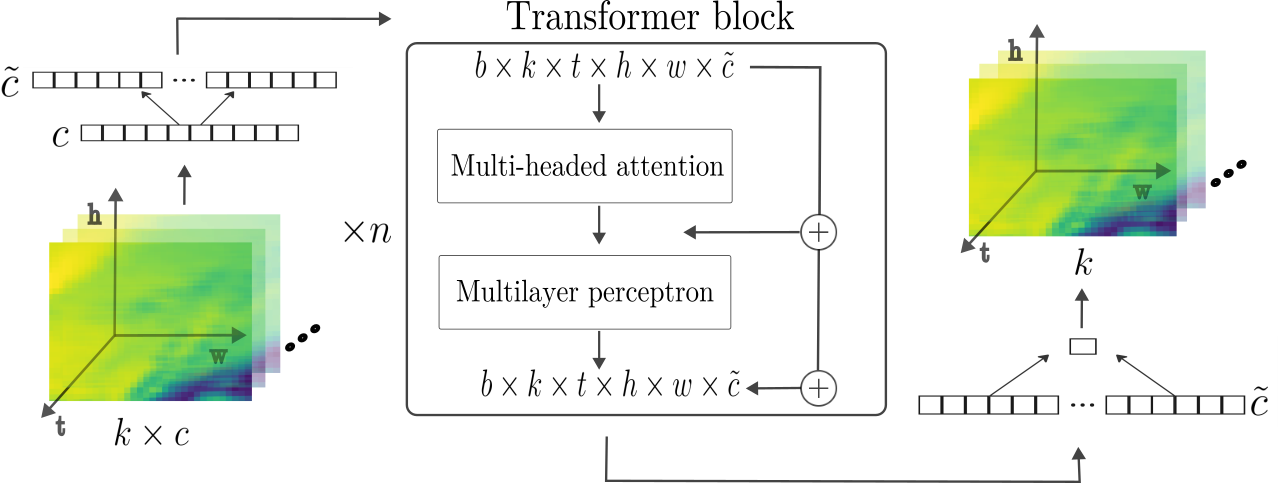}
	\caption{The general architecture of the Transformer-based postprocessing model. The initial data object on the left is a tensor containing $k$ ensemble members with forecasts of $c$ predictors for $t$ lead times on a $h \times w$ latitude-longitude grid. The $c$ predictors are projected towards $\tilde{c}$ features, before passing $n$ times through a transformer block, which contains a block built around self-attention and a multilayer perceptron block. Lastly, the feature dimension is projected back to size one, representing the postprocessed variable.}
	\label{fig:architecture}
\end{figure*}

Transformers are a class of neural networks, initially developed by \cite{Vaswani} to overcome the computational constraints of recurrent neural networks. Their model revolutionized the world of natural language processing due to its significant capacity for parallelization while still finding meaningful dependencies between input and output. At the core of these transformers lies the attention mechanism, a function which is capable of capturing intricate dependencies across multiple dimensions. This is particularly useful for the postprocessing of ensemble weather forecasts, where a variety of meaningful relationships can exist across spatial, temporal, variable and ensemble dimensions \citep{Leutbecher}. Transformers have become widely popular over various scientific domains, in addition to natural language processing, such as computer vision \citep{Vision}, bioinformatics \citep{Bio} and drug discovery \citep{Drug}, and have become synonymous with fast and accurate performance. As described in the previous section, this architecture outperforms other methods for most variables when it comes to statistical postprocessing. The architecture employed here is a modern adaptation of the initial ensemble Transformer developed by \cite{Finn}, adapted to postprocess multiple variables at all lead times at once. The overall architecture of the model is depicted in Figure~\ref{fig:architecture}.
The data is batched as a tensor $\mathbf{Z} \in \mathbb{R}^{b \times k \times t \times h \times w \times c}$, where $b$ stands for the batch size, $k$ represents the number of ensemble members, $t$ the lead times, $h$ the latitude, $w$ the longitude and $c$ the number of predictors. This tensor is first passed through a linear projection where the $c$ predictors are mapped to a higher-dimensional feature space $\tilde{c}$, after which $n$ transformer blocks follow. Finally, another linear layer projects the feature dimension to size one, representing the postprocessed variable. Each transformer block consists of two core components: a block built around multi-headed self-attention and a feed-forward multilayer perceptron (MLP) layer, which are explained in more detail in the following subsections.

\subsubsection{Attention block}
Flowing into the attention block, the input tensor is first normalized across the channel (or feature) dimension $\tilde{c}$. The normalized tensor $\mathbf{Z}_n^l$, passing through the $l$-th block with $l \in \{1, \dots, n\}$, resides in $\mathbb{R}^{b \times k \times t \times h \times w \times \tilde{c}}$ and is projected by three weight matrices $\mathbf{W}$ towards query, key and value matrices residing in the same feature space:
\begin{equation} \label{eq:matrices}
\begin{cases}
\mathbf{V} = \mathbf{W}_v^l \mathbf{Z}_n^l, \\
\mathbf{K} = \mathbf{W}_k^l \mathbf{Z}_n^l, \\
\mathbf{Q} = \mathbf{W}_q^l \mathbf{Z}_n^l.
\end{cases}
\end{equation}
The matrices $\mathbf{W}$ are estimated separately for the value, key and query transformations and contain globally shared weights across the ensemble, temporal and spatial dimensions. In Eq.~\eqref{eq:matrices}, $\mathbf{V}$ is the value matrix, representing the information to be updated, the query matrix $\mathbf{Q}$ stands for the searched information, and the key matrix $\mathbf{K}$, in a holistic view, represents the answers to the query. The features are subsequently divided over multiple heads. Suppose the number of heads equals $h_n$, then each head attends to $\tilde{c}_n = \tilde{c} / h_n$ features, extracting information from a different part of the feature space. At this point, the temporal, spatial and feature dimensions are flattened together, and the matrices $\mathbf{V}$, $\mathbf{K}$ and $\mathbf{Q}$ now reside in $\mathbb{R}^{b \times h_n \times k \times (t \cdot h \cdot w \cdot \tilde{c}_n)}$.

Next, the key and query matrices are normalized and scaled by a factor $s_c = 1 / \sqrt{t \cdot h \cdot w \cdot \tilde{c}_n}$. The attention scores are then computed as the dot product:
\begin{equation}
\mathbf{A} = \text{Softmax}(\mathbf{Q}_n \mathbf{K}_n^\top) \cdot \mathbf{V}, \label{eq:attention}
\end{equation}
where $\mathbf{Q}_n$ and $\mathbf{K}_n$ represent the normalized versions of $\mathbf{Q}$ and $\mathbf{K}$. The output $\mathbf{A}$ is reshaped back to dimensions $\mathbb{R}^{b \times k \times t \times h \times w \times \tilde{c}}$. Although self-attention is applied only across the ensemble dimension, the use of globally shared attention weights $\text{Softmax}(\mathbf{Q}_n \mathbf{K}_n^\top)$ allows information to be implicitly transferred across space, time, and features, similar to the ensemble Kalman filter. This information flow is further discussed in Appendix A. 
Finally, a linear projection with weights $\mathbf{W}_O^l$ precedes the residual connection:
\begin{equation}
\mathbf{Z}_O^l = \mathbf{Z}^l + \mathbf{W}_O^l \mathbf{A},
\end{equation}
resulting in the output tensor $\mathbf{Z}_O^l \in \mathbb{R}^{b \times k \times t \times h \times w \times \tilde{c}}$.
\subsubsection{MLP block}
After the attention function is applied, the tensor flows through an MLP block. It first passes through a normalization layer, after which the feature dimension $\tilde{c}$ is inflated by a multiplication factor $m_n = 4$, allowing the network to model higher-order interactions. This is followed by a Gaussian Error Linear Unit (GeLU) activation function \citep{GeLU}. Another linear layer then projects the feature dimension back to $\tilde{c}$. Finally, a residual connection adds the MLP’s output back to its input, completing the transformer block. 
 
\subsection{Benchmark: Classical member-by-member approach}
As a benchmark method, we apply a classical, yet well-performing, statistical postprocessing method which is referred to in the remainder of this work as classical MBM and is described in detail by \cite{VanSchaey}. The method is often referred to as simply `MBM' in the literature, but given that any method which corrects ensemble members individually technically falls in the member-by-member class, we opt to call it `classical MBM'. \\~\\
The method corrects each member of an ensemble individually according to Eq. \eqref{eq:MBM}:
\begin{equation}
Z^m_C = \alpha + \sum_{i=1}^c \beta_i \bar{V}_i + \tau \epsilon^m \label{eq:MBM},
\end{equation}  
Where $Z^m_C$ represents the corrected value of the postprocessed variable for ensemble member $m$. In this equation, $\alpha$ and $\beta_i$ are regression parameters equal for all members, where the former represents a simple bias parameter and the latter applies a correction proportional to the mean of predictor $i$, i.e., $\bar{V_i}$ with $i \in \{1,..., c \}$ for a total of $c$ predictors. The spread of the ensemble is nudged by the parameter $\tau$ which applies a correction proportional to the difference $\epsilon^m = V^m - \bar{V}$ between the value of the ensemble member $m$ and ensemble mean for the postprocessed variable in question. This parameter $\tau$ represents both an additive and multiplicative correction to the ensemble spread through its dependence on two extra estimated parameters $\gamma_1$ and $\gamma_2$:
\begin{equation}
\tau^2 = \gamma_1^2 + \gamma_2^2 \sigma_{\epsilon}^{-2} \label{eq:tau},
\end{equation}  
where $\sigma^2_{\epsilon}$ represents the ensemble variance. While the first two terms of Eq. \eqref{eq:MBM} and $\gamma_1$ and $\gamma_2$ of Eq. \eqref{eq:tau} are shared among all members, the last term of Eq. \eqref{eq:MBM} is unique to every member through its dependence on $\epsilon^m$. While   $\alpha$ and $\beta_i$ are shared across members at each grid point, it is important to note that all parameters of classical MBM are estimated separately for each grid point and lead time, resulting in a location-specific correction. The original framework developed by \cite{VanSchaey} provides multiple approaches for estimating the parameters in Eq. \eqref{eq:MBM} and \eqref{eq:tau}, with varying complexity and computational expense, depending on the type of reliability constraints that are enforced. The application of classical MBM techniques is facilitated by the \textit{pythie} software, as developed by \cite{pythie}.    

  \subsection{Data}
\begin{table}[]
	\centering
	\caption{All parameters utilized as predictors in the experiments described in Section \ref{sec:results} are listed below. All variables are surface variables, except for the last six. Orography is a static variable, while Geopotential height at 500 hPa, wind speed at 700 hPa, temperature at 850 hPa and U and V wind components at 700 hPa are level variables.}
	\label{table:predictors}
	\begin{tabular}{@{}llll@{}}
		\toprule
		\textbf{Parameter name}    & \textbf{Short name} & \textbf{Units} & \textbf{Predictor for} \\ \midrule
		2 m temperature            & t2m                 & $^{\circ}$C              & t2m, w10 and w100        \\
		10 m U wind component      & u10                 & ms$^{-1}$             & t2m, w10 and w100         \\
		10 m V wind component      & v10                 & ms$^{-1}$             & t2m, w10 and w100         \\
		10 m wind speed            & w10                 & ms$^{-1}$             & t2m, w10 and w100         \\
		10 m wind gusts			  & p10fg6			   & ms$^{-1}$             & t2m, w10 and w100         \\
		100 m wind speed  		  & w100  			   & ms$^{-1}$             & w100         \\
		100 m U wind component     & u100                & ms$^{-1}$             & w100         \\
		100 m V wind component     & v100                & ms$^{-1}$             & w100         \\
		Total cloud cover          & tcc                 & $ \in [0,1]$        & t2m, w10 and w100         \\
		Snow depth                 & sd                  & m              & t2m           \\
		Maximum temperature at 2 m & mx2t6               & $^{\circ}$C               & t2m           \\
		Minimum temperature at 2 m & mn2t6               & $^{\circ}$C               & t2m        \\
		Geopotential height at 500hPa        & z                   & m$^2$s$^{-2}$             & t2m, w10, w100       \\
		Wind speed at 700 hPa  	  & w700  			   & ms$^{-1}$             & w100         \\
		Temperature at 850 hPa     & t                   & $^{\circ}$C         & t2m, w10, w100 \\
		U wind at 700 hPa          & u                   & ms$^{-1}$         & w100         \\
		V wind at 700 hPa          & v                   & ms$^{-1}$         & w100         \\
		Orography                  & oro                 & m              & t2m, w10, w100          \\ 
		\bottomrule
	\end{tabular}
\end{table}

The performance of complex deep learning algorithms depends heavily on the data they are trained upon, as their predictive power is limited by the information present in that data \citep{Dueben}. Comparing postprocessing methods trained on different datasets is therefore not straightforward. Inspired by this difficulty, \cite{Demaeyer} developed a benchmark dataset, the EUPPBenchmark dataset, with the comparison of various postprocessing methods as the main goal\footnote{The complete dataset is publicly available and can be downloaded at https://github.com/EUPP-benchmark/climetlab-eumetnet-postprocessing-benchmark \citep{Demaeyer}.}. The dataset contains forecasts and reforecasts, i.e., reruns of the current NWP models with historical initial conditions, generated by the Integrated Forecast System (IFS) of the European Centre for Medium-Range Weather Forecasts (ECMWF), and spans the period from 1997 until 2018. The dataset covers significant parts of West and Central-Europe, as depicted in Figure \ref{fig:area}, on a $0.25^{\circ} \times 0.25^{{\circ}} $ grid, corresponding to a resolution of roughly 25 km. This resolution is in alignment with the ERA5 reanalysis dataset, which is included in the benchmark dataset as ground truth \citep{Hersbach}. To avoid data leakage between training and testing phases, the data was split chronologically: 1997–2015 for training, 2016 for validation, and 2017 for testing. On each reforecast date, one ensemble forecast containing eleven members is initialized at 00:00 UTC, with forecasts ranging from a lead time of 0 up until 120 hours in the future. The EUPPBench dataset contains a myriad of meteorological variables, both at the surface and on various pressure levels. Table 1 contains the subset of these variables employed as predictors for the postprocessing of both temperature (t2m) and wind speed at ten meter (w10) and one hundred meter (w100). 
\subsection{Methodological Approach}
\begin{figure}[H]
	\centering
	\includegraphics[scale=0.45]{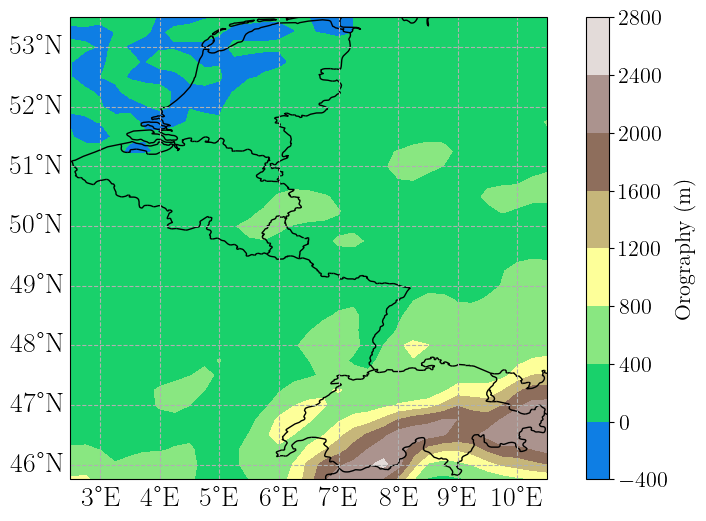}
	\caption{The area, including Belgium, the Netherlands, Luxembourg, Switzerland and large parts of Germany and France, covered by the data in the EUPBBench dataset.}
	\label{fig:area}
\end{figure}

Temperature, ten-meter and one hundred-meter wind speed  forecasts covering the entire EUPPBench  domain depicted in Figure \ref{fig:area}  are postprocessed by the Transformer and classical MBM using the meteorological variables presented in Table \ref{table:predictors} as predictors. It should be noted that, in this work, the postprocessing methods are applied to gridded forecasts instead of the station-level data employed in the original setup of \cite{Demaeyer}. Given that classical MBM is often implemented with only one predictor, such as for example by \cite{Demaeyer}, we initially tested classical MBM using only the target variable as input, and subsequently evaluated the method using the same set of predictors as the Transformer model. Both versions were tested across a selection of lead times. For temperature, the multi-predictor variant outperformed the single-predictor version and was therefore used, only orography was excluded for classical MBM, because the static field led to the occurrence of singular matrices during the minimization process of the \textit{pythie} package. Further evaluation revealed that the inclusion or exclusion of orography as a predictor did not lead to statistically significant improvements in model performance for the Transformer.  In contrast, for wind speed, both at ten and one hundred meter, the single-predictor variant of classical MBM yielded slightly better performance while also being considerably more efficient computationally. For each target variable, we report results for the most competitive MBM variant. All regression parameters where estimated by minimizing the Continuous Ranked Probability Score (CRPS), in essense the probabilistic version of the mean absolute error. 
\begin{table*}[ht]
	\centering
	\caption{Training times (in hours) for classical MBM and the Transformer for two-meter temperature, ten-meter wind speed and one hundred-meter wind speed.}
	\label{table:training_times}
	\begin{tabular}{@{}lccc@{}}
		\toprule
		\textbf{Method}                  & \textbf{T2M (hours)} & \textbf{W10 (hours)} & \textbf{W100 (hours)} \\ 
		\midrule
		Classical MBM                    & $\sim$ 9.4           & $\sim$ 3.3           & $\sim$ 2.3            \\
		Transformer                      & $\sim$ 1.4           & $\sim$ 0.6           & $\sim$ 0.5            \\ 
		\bottomrule
	\end{tabular}
\end{table*}

This score quantifies the area between the predictive, postprocessed distribution of the ensemble members and the actual observation:
\begin{equation}
\text{CRPS}(F,z) = \int_{\mathbb{R}} \Big( F(y) - \mathbb{I}\{z \leq y\}\Big)^2,		\label{eq:CRPS}
\end{equation}     
where $F$ is the cumulative distribution function, and $\mathbb{I}$ the indicator function. For temperature, we follow previous work in assuming a normal predictive distribution \citep{Wilks, Finn, Ashkboos, Bouallegue}, resulting in the possibility to express Eq. \eqref{eq:CRPS} analytically using the normal cumulative distribution and probability density functions \citep{Gneiting05}, an expression that is given explictly in the Appendix (Eq. \eqref{eq:CRPSnormal}). Classical MBM was trained in a similar way, i.e. with the class \textit{EnsembleSpreadScalingNgrCRPSCorrection} of the \textit{pythie} package, which minimizes the CRPS assuming a normal distribution. This distributional assumption was used both for training and for verification. For wind speed, however, the assumption of normality is less appropriate, as it allows for negative values and may not reflect the true shape of wind speed distributions, given the variety of distributions already proposed for wind speed \citep{JUNG}. To avoid introducing an ill-fitting assumption, we opted for distribution-free methods. During training, the Transformer was optimized using a regularized, kernel-based CRPS loss based on the expression given by \cite{Gneiting2}. For MBM, we used the \textit{EnsembleAbsCRPSCorrection} class from the \textit{pythie} package, which also minimizes the CRPS directly while enforcing non-negativity. In order to verify the postprocessed wind speed forecasts, we employed the fair ensemble CRPS of \cite{Leutbecher2} for both models. All the CRPS expressions are provided explicitly in Appendix B.
\\~\\
The Transformer was trained in batches of 2 samples, using a learning rate of 0.001 with Adam optimizer \citep{Adam}. To prevent overfitting, weights were only saved when improvement in the loss functions was observed  and early stopping was employed, i.e. training was terminated when there was no improvement for five consecutive epochs. The code employed is an adaptation from scripts made available by \cite{Finn} and \cite{Ashkboos}. The attention module described in Subsection \ref{subsec:Transformer} of Section \ref{sec:methods} was repeated four times ($n=4$), with eight different attention channels or heads, i.e. $h_n=8$ and a multiplication factor $m_n=4$. Training time per lead time was $27.5\%$ shorter when compared to training the Transformer separately per lead time and inclusion of ten predictors only prolonged training time by $7\%$ compared to three predictors, and by $12 \%$ compared to one predictor. Training was carried out on a workstation employing two graphics cards (MSI 24GB D6X RTX 4090 Gaming X Trio). Regressing classical MBM took, on the other hand, more than six times longer for temperature and around five times longer for wind speed at ten and one hundred meter, respectively, with approximate training times presented in Table \ref{table:training_times}. Training times are, however, mostly indicative, as the Transformer benefits from the GPUs, while classical MBM relies on the parallelization of computations across CPUs, making an exact comparison difficult. Lastly, it is important to note that for the wind variables, the Transformer was trained on substantially more data, as a vast amount of predictors were included as compared to the single predictor for classical MBM. \\~\\
In order to assess the quality of the postprocessed forecasts, a variety of verification scores are computed. 
First, we calculate the CRPS and bias of the postprocessed test data on average over all lead times, on average over the spatial grid and lastly averaged over all lead times and spatial dimensions. These scores reflect the overall accuracy of the postprocessed ensembles but, however, reveal little information regarding their variability. Given that countering underdispersiveness, which leads to overconfident models, is one of the central goals of postprocessing \citep{Lakatos}, assessing scores quantifing uncertainty is essential. Consequently, we calculated the ensemble spread and the spread-error ratio (SER), where the latter is obtained by dividing the former by the root mean squared error (RMSE). Following \cite{Fortin}, the spread is calculated by taking the square root of the averaged variance instead of just averaging out the standard deviation of the ensemble forecasts. Ideally, the spread of an ensemble matches its error, resulting in a well calibrated ensemble where the uncertainty of the forecast is well reflected in the ensemble \citep{Scher}, as such, a SER value close to one is desired. Next, we assess the reliability of the postprocessed ensemble forecasts by calculating the rank histograms. In a perfectly reliable ensemble, the observation has equal probability to fall between any two members, meaning that it is equally probable that the observation has rank $i$ (falling between observation $i-1$ and $i$) as rank $k$ (falling between observation $k-1$ and $k$). Consequently, a perfect reliable ensemble results in a uniform histogram \citep{Keller}. \\~\\
Lastly, we investigate the influence per geographic region by probing the heart of the correction applied by the Transformer, i.e., the application of self-attention in Eq. \eqref{eq:attention}, in the element-wise product $\mathbf{Q} \odot \mathbf{K}$. Following \cite{Finn}, attentive regions with high influence can be unveiled by averaging the element-wise product of the key and value matrix over member and time dimensions, a process that is described in more detail in Appendix A. This calculation is carried out for all channels, at the end of the last iteration trough the attention module, resulting in an attention map for every channel. As such, the magnitude of the impact of every region on the postprocessing process can be assessed.

 \section{Results} \label{sec:results}
\subsection{Temperature}
\begin{table*}[ht]
	\centering
	\caption{CRPS and SER values for all methods, for two-meter temperature, ten-meter wind speed and one hundred-meter wind speed. Values are averaged over all temporal and spatial dimensions. Bold values indicate best performing methods.}
	\label{table:scores}
	\begin{tabular}{@{}lrrrrrrr@{}}
		\toprule
		\multirow{2}{*}{\textbf{Method}} & \multicolumn{2}{c}{\textbf{T2M }} & \multicolumn{2}{c}{\textbf{W10}} & \multicolumn{2}{c}{\textbf{W100}} \\ 
		\cmidrule(lr){2-3} \cmidrule(lr){4-5} \cmidrule(lr){6-7}
		& CRPS {[}K{]} & SER & CRPS {[}m/s{]} & SER & CRPS {[}m/s{]} & SER \\ \midrule
		Raw                             & 1.008       & 0.542         & 0.519     & 0.753         & 0.802       & 0.756         \\
		Classical MBM                   & 0.889       & 0.921         & 0.477      & 0.878        & 0.747      & 0.865         \\
		Transformer                     & \textbf{0.841}  & \textbf{0.986}  & \textbf{0.467}  & \textbf{1.009}    & \textbf{0.732} & \textbf{0.952} \\ \bottomrule
	\end{tabular}
\end{table*}

\begin{figure*}[h!]
	\centering
	\includegraphics[scale=0.675]{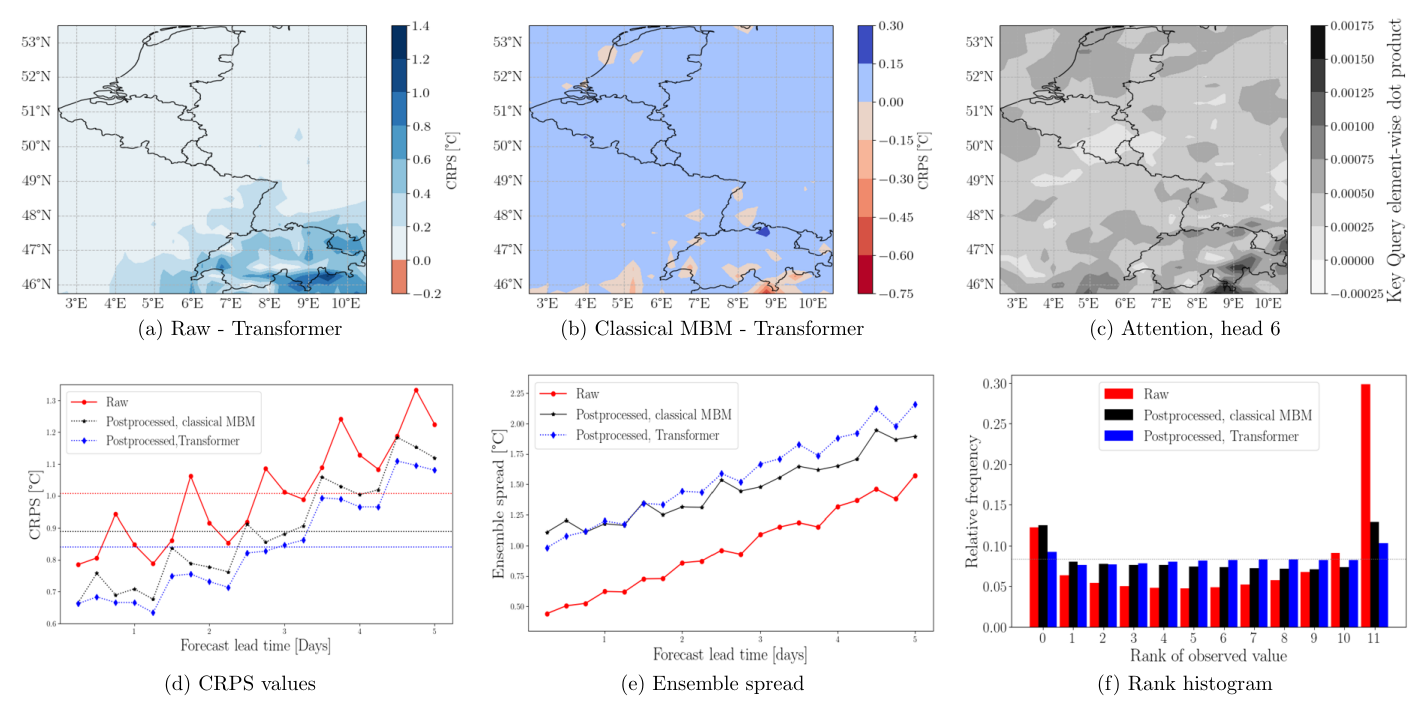}
	\caption{\textit{Top row}: Improvement in CRPS when comparing the postprocessed forecasts of temperature of the Transformer with  (a) the original forecasts and (b) classical MBM. The bluer the region, the better the performance of the Transformer. A map, corresponding to head 6, showing attentive regions of high influence in the last attention module for temperature is shown in Figure (c). A darker color means that particular region had a larger influence over the attention weights. \textit{Bottom row:} Evaluation metrics for the postprocessed forecasts: (d) CRPS in function of lead time (lower is better), (e) Ensemble spread in function of lead time and (f) Rank histogram (uniform is better).}
	\label{fig:AllGraphsTemp}
\end{figure*}

Average CRPS and SER scores are presented in Table \ref{table:scores}, with bold values indicating the best performing method per score, where the Transformer comes out as the best performing method overall. It is important to note that these values are averaged over all dimensions, and a further investigation per location and lead time is required to assess anomalies in the performances. Panel (a) of Figure \ref{fig:AllGraphsTemp} shows the difference in CRPS of the Transformer over the grid as compared to the original forecasts, with in this case up to 1.4 degrees decrease in CRPS in the Alps. The difference over the grid between the two methods is depicted in panel (b), where it can be inferred in what areas the  Transformer improves over classical MBM or vice-versa. The map shows an overall blue colour over most regions of the mainland indicating a better performance of the Transformer, although there are small regions in the South where classical MBM performs better. Lastly, we calculated the CRPS values per lead time, which are shown in Figure \ref{fig:AllGraphsTemp}, panel (d). The Transformer results in the best performance for each lead time: classical MBM realizes an average improvement of $12\%$,  whereas the Transformer improves $16.5\%$ when comparing to the original forecasts. For bias values per lead time, together with the bias gradient over the grid, we refer the reader to panel (a) of Figure \ref{fig:biaslead} and the top row of Figure \ref{fig:biasgrid} in Appendix C. Here, it can be seen that both methods achieve a significant improvement over the bias of the raw forecasts over all lead times, with a more pronounced negative and positive bias of the Transformer for the region around Switzerland, explaining the better performance of classical MBM in that region, as shown in the top row of Figure \ref{fig:biasgrid}. Next, image (c) of Figure \ref{fig:AllGraphsTemp} shows one of the eight attention \textit{heads} in action, revealing attentive regions with high influence. This figure clearly shows a high influence of the region around the Alps, while focusing much less on other parts of the mainland and the sea.

Lastly, we assess the uncertainty and reliability of the forecasts. Panel (e) of Figure \ref{fig:AllGraphsTemp} reveals a significant enlargement in ensemble spread for both classical MBM and the Transformer, where the latter grows more steep with lead time as compared to the former. This is also reflected in panel (a) of Figure \ref{fig:reslead} in the Appendix which shows that both the Transformer and classical MBM significantly improve on the SER and result in values close to one per lead time. The rank histograms are shown in panel (f) where it can be seen that the original ensemble is largely overconfident, resulting in an underdispersed, i.e. not enough spread, $\bigcup$-shaped histogram \citep{Thorar}. Classical MBM and the Transformer result in a much more uniform distributed rank histogram, although the probability for the observation to fall under the smallest or above the largest members remains too large for classical MBM, while being smaller for the Transformer.

\subsection{ten-meter wind speed} \label{sec:wind}

\begin{figure*}[h!]
	\centering
	\includegraphics[scale=0.675]{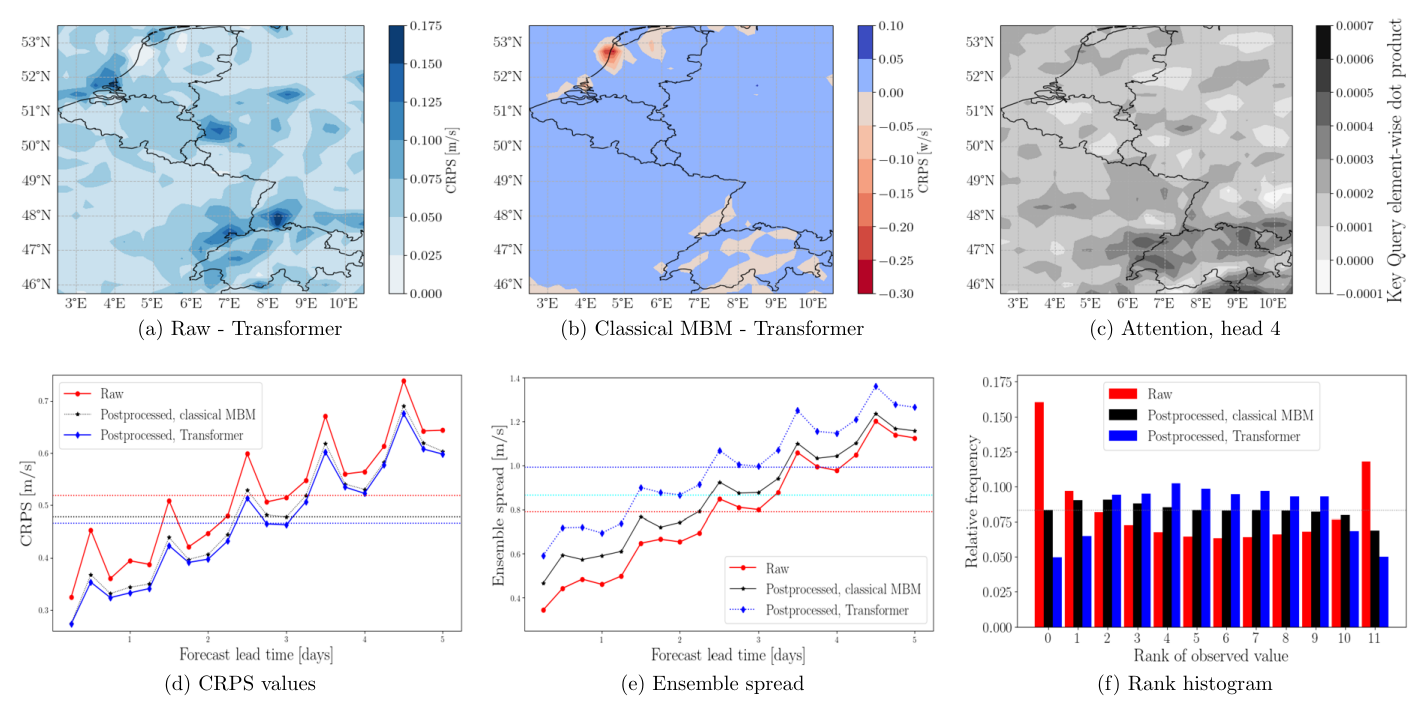}
	\caption{\textit{Top row}: Improvement in CRPS when comparing the postprocessed forecasts of ten-meter wind speed of the Transformer with  (a) the original forecasts and (b) classical MBM. The bluer the region, the better the performance of the Transformer. A map, corresponding to head 4, showing attentive regions of high influence in the last attention module for ten-meter wind speed is shown in Figure (c). A darker color means that particular region had a larger influence over the attention weights. \textit{Bottom row:} Evaluation metrics for the postprocessed forecasts: (d) CRPS in function of lead time (lower is better), (e) Ensemble spread in function of lead time and (f) Rank histogram (uniform is better).}
	\label{fig:AllGraphs10mwind}
\end{figure*}

The results for ten-meter wind speed follow a similar trend as for temperature: concerning the CRPS, the Transformer outperforms classical MBM over most regions of the EUPPBenchmark dataset area and at all lead times, as shown in panels (b) and (d) of Figure \ref{fig:AllGraphs10mwind}. Regarding the overall CRPS score, the Transformer improves with  $10\%$ on the raw forecasts, as compared to $8\%$ for classical MBM. Performance is, however, worse around small regions in the Netherlands and some areas in Switzerland when comparing to classical MBM. Interestingly, these areas are marked as attentive regions with high influence on the attention map, i.e. subfigure (c) of Figure \ref{fig:AllGraphs10mwind}. Next, when calculating the bias of the postprocessed wind speed forecasts, it becomes apparent that the Transformer results in a slightly negative bias over all lead times (picture (c) of Figure \ref{fig:biaslead}). The reason behind this underestimation becomes clear when examining the average bias induced by the Transformer over the domain, averaged over the lead times, as shown in the second row of Figure \ref{fig:biasgrid}. Overall, bias is minimal and varies little but smoothly over the spatial grid, except for the North-West of the Netherlands, where a strong negative bias appears which results in a negative bias for larger lead times. Next, when assessing the metrics quantifying the uncertainty of the postprocessed ensemble, differences among methods become more apparent. Panel (e) of Figure \ref{fig:AllGraphs10mwind}, for example, shows a steady increase in ensemble spread for the Transformer, while the rate of growth of the spread seems to drop with lead time for classical MBM, coinciding with the underdispersed raw forecasts at the largest lead times. This result is naturally also reflected in the SER, depicted in panel (b) of Figure \ref{fig:reslead}, showing almost perfec SER values for the Transformer as opposed to an overconfident benchmark method which again almost matches with the raw ensemble forecasts at larger lead times. Lastly, the rank histograms of both the Transformer and classical MBM both improve on the initially overconfident raw ensemble, as visible in panel (f) of Figure \ref{fig:AllGraphs10mwind}. The histogram of the classical MBM shows a slight right skewness, indicating a small overestimation of the observed wind speed, while the Transformer results in an overdispersed rank histogram, indicating that observations have a lower chance of falling outside the ensemble's lower and upper bounds.      

\subsection{One hundred-meter wind speed} \label{sec:100w}
\begin{figure*}[h!]
	\centering
	\includegraphics[scale=0.675]{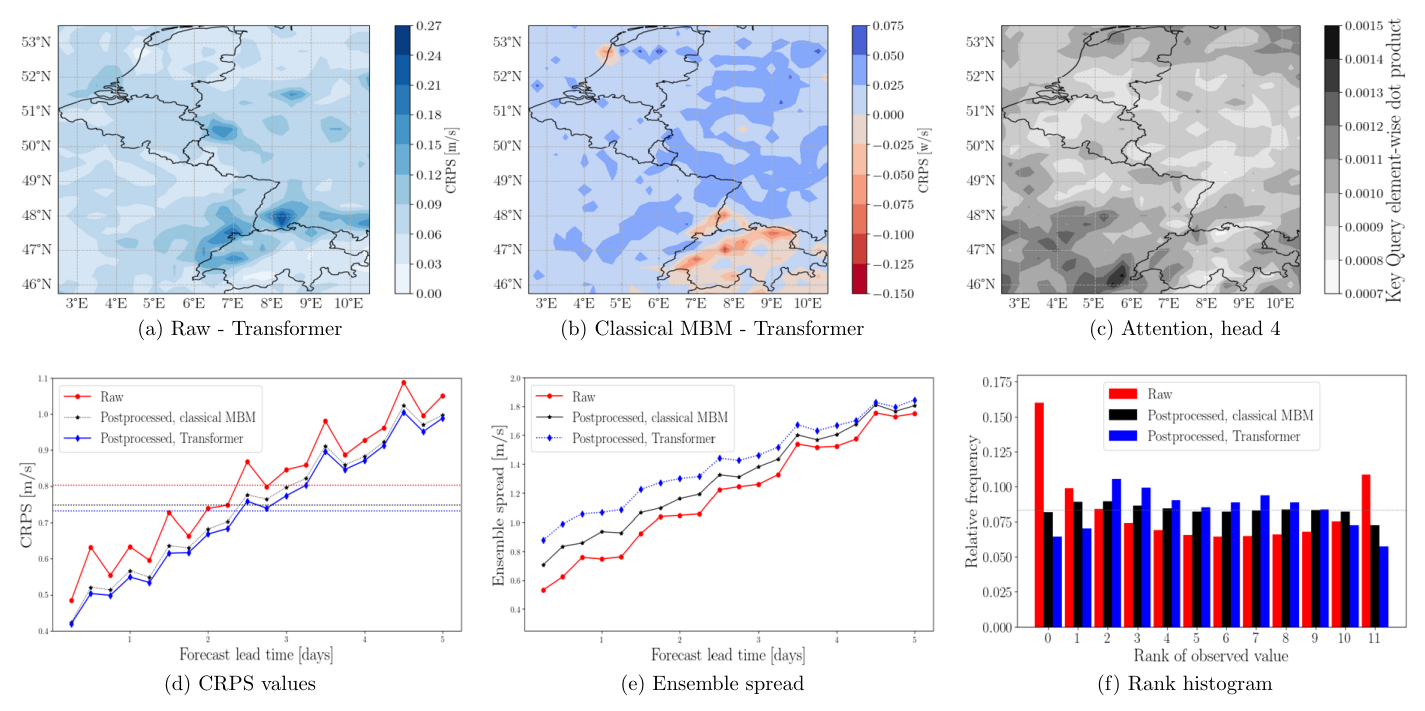}
	\caption{\textit{Top row}: Improvement in CRPS when comparing the postprocessed forecasts of one hundred-meter wind speed of the Transformer with  (a) the original forecasts and (b) classical MBM. The bluer the region, the better the performance of the Transformer. A map, corresponding to head 4, showing attentive regions of high influence in the last attention module for one hundred-meter wind speed is shown in Figure (c). A darker color means that particular region had a larger influence over the attention weights. \textit{Bottom row:} Evaluation metrics for the postprocessed forecasts : (d) CRPS in function of lead time (lower is better), (e) Ensemble spread in function of lead time and (f) Rank histogram (uniform is better).}
	\label{fig:AllGraphs100mwind}
\end{figure*}

Concerning one hundred-meter wind speed, results are more mixed as compared to temperature and ten-meter wind speed. Panel (a) of Figure \ref{fig:AllGraphs100mwind} shows that the Transformer realizes a significant improvement in CRPS over the spatial grid, especially in the South-East region of the EUPP Benchmark. When comparing with the benchmark method, as depicted by panels (b) and (d), it becomes clear the Transformer performs better across the vast region of the map, except for Switzerland and a very small region in the Netherlands. The Transformer performs better across all lead times with a general improvement of $9\%$ as compared to $7\%$ for the benchmark method. When probing the attentive regions with high influence, as shown by subfigure (c) of Figure \ref{fig:AllGraphs100mwind}, large relative differences between various regions arise. In descending order of importance, the attention channel appears to assign larger weights to the South-West, the Alps region, and finally the vast mainland region. Bias values for both classical MBM and the Transformer, both per lead time and over the grid, are  similar to those of ten-meter wind speed. The Transformer results in an increasingly negative bias for larger lead times, with a sharp negative bias in some parts of the North and a more positive value, generally speaking, in the Southern area, as depicted in panel (c) of Figure \ref{fig:biaslead} and the bottom row of Figure \ref{fig:biasgrid} in Appendix C. When assessing the uncertainty metrics, panel (e) shows that the Transformer first results in a steady increase with lead time for the ensemble spread, after which the increase in spread of the ensemble postprocessed by both the Transformer and Classical MBM tends to stagnate and end on nearly equal foot with the spread of the original forecasts. SER values, depicted in panel (c) of Figure \ref{fig:reslead}, show an initially well balanced ensemble for the Transformer, which becomes underdispersed  with longer lead times, wheres the classical MBM provides an ensemble which is overconfident over all lead times. Lastly, the rank histograms show overconfident raw forecasts, a relatively uniform, slightly right-skewed histogram for classical MBM and a slightly underconfident ensemble for the Transformer.     

\section{Discussion and future prospects} \label{sec:conclusions} 
This work proposes a Transformer for the postprocessing of ensemble weather forecasts, effectively realizing a very fast, accurate technique which easily allows the inclusion or exclusion of multiple lead times, predictors and ensemble members at low computational cost. Performance is compared with the classical, yet well-performing MBM method, for two-meter temperature and wind speed, both at ten and one hundred meter, over the complete domain of the EUPPBenchmark dataset. The Transformer improves on the raw forecasts with $16.5\%$ for temperature, $10\%$ for ten-meter wind speed and $9\%$ for one hundred-meter wind speed. Overall, performance of the Transformer is better than that of classical MBM for all lead times and the mainland of the EUPPBench dataset. When comparing training time, the Transformer requires around six times less training time for temperature and five times less for ten-meter wind speed and one hundred-meter wind speed. Next, the Transformer significantly improves the spread over the lead times as compared to both the raw forecasts and classical MBM, where differences are more pronounced for wind speed than for temperature. Both postprocessing methods produce relatively well-calibrated ensembles. The Transformer shows consistently better spread-error ratio (SER) values, while classical MBM tends to be underdispersed in this regard. However, for wind speed, the rank histograms of MBM are more uniform, compared to the Transformer, which shows signs of overdispersion.  \\~\\  
Concerning accuracy per region, as measured by the CRPS, performance of the Transformer is heterogeneous over the EUPPBench area. Around the North Sea, for example, improvements for one hundred-meter wind speed are significant in some regions, which is a relevant result in the light of renewable energy production, given that a major part of wind energy is generated offshore \citep{Rumes, Sorensen}. Performance of the Transformer for wind speed appears to be worse, however, in very small regions in the Northwest of the Netherlands, i.e. the Amsterdam area and Zeeland on panels (b) of Figures \ref{fig:AllGraphs10mwind} and \ref{fig:AllGraphs100mwind}, even though these region share a similar topography with their direct surroundings. The reason for this might be a mismatch in near surface winds between assimilation cycles in the ERA5 data, which is known to lead to inaccuracies in capturing coastal wind dynamics \citep{ECMWF}. Around Switzerland, in the Alps region, however, there are some regions where the postprocessed forecasts are less accurate than the benchmark method, both for temperature and wind speed. This can partly be explained by the fact that classical MBM is applied locally, while the Transformer models spatiotemporal dependencies more globally. Further investigation reveals that those regions have a relatively large influence on the residual connection of the attention module through the key and value tensors, both for temperature and wind speed. This can be  most likely explained by higher CRPS values in this region during training, which naturally gives those grid points greater influence on the loss function. Next, when assessing the bias over the domain for wind speed, the heterogeneity of the magnitude of the bias of the Transformer over the shores of Belgium and the Netherlands is remarkable when comparing to the benchmark method, whereas this phenomenon can be observed in the Alps region for temperature.  Given the regional dependence of the quality of the performance of the Transformer, a promising avenue for future work arises: dividing the grid points of Figure \ref{fig:area} into clusters based on meteorological characteristics. This could for example be realized by performing cluster analysis on the benchmark dataset with respect to multiple atmospheric variables and consequently obtaining a classification based on different synoptic weather types, similar to work by \cite{Arroyo}. As such, an attention-based model could be trained with different weights for each cluster and the weights would become region-dependent, resulting in a more specialized and local postprocessing approach, which might help improve the overall performance of the Transformer. Additionally, as suggested by \cite{Bouallegue}, a combination of a Transformer with a more classical method could be possible to attain the best performance in every region. This approach may be particularly effective since the classical MBM method still surpasses the Transformer in a small portion of the test area. Further investigation could help uncover the reasons behind this disparity and facilitate the development of an optimized combination strategy. Besides that, follow-up research constituting of an in-depth analysis comparing the added value of each predictor utilized here for the postprocessing of both temperature and wind speed will be carried out. \\~\\
Interestingly, the Transformer improves significantly when compared to classical MBM for two-meter temperature, while improvements for ten-meter and one hundred-meter wind speed are smaller. A possible explanation could be that the Transformer extracts more relevant information from the set of predictors for temperature, which it might find to be more difficult for wind speed, especially at higher altitudes, given that wind speed relies much more on complex, vertical processes, like e.g. turbulence \citep{Monahan}. Another contributing factor might be the quality of the reanalysis data used as reference: although ERA5 generally performs very well, some studies report that it underestimates strong wind events offshore and struggles to capture variability at coastal regions \citep{Wes, Alkhalidi}, reflecting the need for large, high-quality, gridded reference datasets for wind speed.
\\~\\
Lastly, another important contemplation concerns the future of explicit statistical postprocessing techniques, like the ones presented in this work. As NWP models still constitute the backbone of operational weather forecasting today, these techniques remain necessary in order to improve predictions. Every scientific domain, however, steers away from physics-based, white-box modeling towards data-driven methods, to which weather forecasting forms no exception. \cite{Schultz} raised the question whether deep learning could beat NWP in the near future, and the advent of models like Graphcast \citep{Lam}, Pangu-weather \citep{Bipangu} and Fourcastnet \citep{Pathak} seem to answer that question in favor of the former. The best-performing postprocessing methods today, such as the Transformer discussed here, are entirely data-driven. Consequently, one could expect these methods to be fully integrated into the forecasting model itself, as the need for postprocessing arises from biases and parameterization errors inherent in the NWP model but absent in a purely data-driven approach. Completely data-driven weather forecasting, however, still has drawbacks \citep{Ben} and statistical postprocessing as such continues to be an essential block in the weather forecasting chain and its many downstream applications. Therefore, research with regards to new, or the improvement of existing, techniques remains essential and should be continued.        
\section{Acknowledgments}
The authors extend their thanks to the Belgian Science Policy Office (BELSPO) for providing financial support for this work (B2/223/P1/E-TREND \& Prf-2020-025\_AIM). TSF acknowledges the support of the project SASIP funded by Schmidt Sciences (Grant number G-24-66154) — a philanthropic initiative that seeks to improve societal outcomes through the development of emerging science and technologies. 
\section{Data and code availability }
The EUPPBenchmark dataset used for training and testing can be downloaded at \url{https://github.com/EUPP-benchmark/climetlab-eumetnet-postprocessing-benchmark}. The implementation of the Transformer will be made available at \url{https://github.com/UAntwerpM4S/PP\_EUPP}.  

\appendix
\renewcommand{\thesection}{\Alph{section}} 
\renewcommand{\theequation}{A.\arabic{equation}}
\setcounter{equation}{0} 
\renewcommand{\thefigure}{A.\arabic{figure}}
\setcounter{figure}{0} 
\renewcommand{\thetable}{A.\arabic{table}}
\setcounter{table}{0} 

\newpage
\section{Information flow in the Transformer} \label{appendix:A}

Although the attention mechanism computes interactions explicitly across the ensemble dimension at fixed spatiotemporal coordinates, the architecture allows for implicit information flow across time and space by the shared global attention weights, $\text{Softmax}(\mathbf{Q_n}(\mathbf{K_n})^T)$, in Eq. \eqref{eq:attention}. Additionally, the weights for the queries, keys and values are shared across all grid points. As such, they learn to encode spatiotemporal context into each embedding vector. This implies that each query-key interaction implicitly reflects spatial and temporal correlations learned through the projection matrices. Such behavior is analogous to ensemble Kalman filters (EnKFs), where correlations between spatial and temporal features are captured through ensemble statistics even though the update equations apply locally: the queries are analogous to the observations, the keys to the ensemble members in observational space and the values to the ensemble members in state space. Furthermore, \cite{Choro} showed that the attention function \eqref{eq:attention} can be approximated using low-rank kernel factorization as follows: 
\begin{equation}
\mathbf{A} \approx \phi(\mathbf{Q}) \Big( \phi(\mathbf{K})^T \mathbf{V} \Big),  \label{eq:approx}
\end{equation}
where the regular softmax is approximated with feature maps $\phi$ which map $\mathbf{K}$ and $\mathbf{Q}$ to a lower rank approximation $\phi(\mathbf{K}), \phi(\mathbf{Q}) \in \mathbb{R}^{b \times h_n \times k \times (t_r \cdot h_r \cdot w_r \cdot \tilde{c}_{n,r}))}$. Consequently, the product $\phi(\mathbf{K})^T \mathbf{V}$, over the ensemble dimension, resides in $ \mathbb{R}^{b \times h_n \times (t_r \cdot h_r \cdot w_r \cdot \tilde{c}_{n,r})\times(t \cdot h \cdot h \cdot w \cdot \tilde{c}_n)}$ and effectively signifies the information transfer shared across spatiotemporal positions encoded in the key and value projections. As such, even though the softmax is computed only over the ensemble axis, spatiotemporal structure embedded in the learned projections enables broader information propagation across the entire domain.
This was empirically shown by \cite{Rank}, who conducted a sensitivity analysis of the transformer architecture by evaluating the gradient of a single grid point as compared to all other grid points. Here, it was shown that for a small ensemble size ($k=5$) the gradients were noisy, while for larger ensembles ($k=50$) the gradients showed coherent spatially structured patterns. This confirms that the model learns to propagate global spatial-temporal information through ensemble-based attention, where the global mixing improves with ensemble size. \\~\\
To generate the attention maps shown in panel (c) of Figures \ref{fig:AllGraphsTemp}, \ref{fig:AllGraphs10mwind} and \ref{fig:AllGraphs100mwind}, we compute the element-wise product between the projected query and key tensors, i.e., $\mathbf{Q} \odot \mathbf{K}$, and average over ensemble, temporal and feature dimension. A representative head and batch are selected for visualization, yielding a 2D map that highlights regions with strong average query-key alignment over time, which can be interpreted as the most influential regions in the calculation of the attention-based corrections. It is important to note that these images merely show activate regions and do not represent the actual corrections applied to the ensembles.

\section{Continuous Ranked Probability Score} 
\label{appendix:B} 
If the predicted variable $y$ is assumed to be normally distributed, the analytical expression for the CRPS becomes \citep{Gneiting05}:

\begin{equation}
\text{CRPS}(\mu, \sigma; y) = \sigma \left[ \frac{y - \mu}{\sigma} \left(2 \Phi\left( \frac{y - \mu}{\sigma} \right) - 1 \right) + 2 \phi\left( \frac{y - \mu}{\sigma} \right) - \frac{1}{\sqrt{\pi}} \right], \label{eq:CRPSnormal}
\end{equation}

where $\Phi(\cdot)$ is the cumulative distribution function (CDF), $\phi(\cdot)$ is the probability density function (PDF) of the standard normal distribution and $\mu$ and $\sigma$ refer to the mean and the standard deviation of the distribution. This expression was minimized when estimating the parameters of the different methods for the postprocessing of temperature, and when verifying those forecasts. \\~\\
For wind speed, we employed a distribution-free kernel-based CRPS during Transformer training, adapted from the formulation by \citep{Gneiting05} with an added regularization term to penalize excessive spread:
\begin{equation}
\text{CRPS}_{\text{kernel}} = \frac{1}{m} \sum_{i=1}^{m} |x_i - y| - \frac{1}{2 m^2} \sum_{i=1}^{m} \sum_{j=1}^{m} |x_i - x_j| + \lambda \cdot \text{Penalty},
\end{equation}
where $m$ is the number of ensemble members, $x_i $ are the ensemble forecasts, $y$ is the observation, and the regularization term is defined as:

\begin{equation}
\text{Penalty} = \frac{1}{m} \sum_{i=1}^{m} \max\left(0,\ |x_i - \bar{x}| - k \cdot \sigma_x\right),
\end{equation}
with $\bar{x}$ and $\sigma_x$ being the ensemble mean and standard deviation, and \( k \), \( \lambda \) being tunable constants. In our implementation, we used  $\lambda = 0.0275$ and  $k = 2.7$ for ten-meter wind speed, while for one hundred-meter wind speed we used $\lambda = 0.05$ and $k = 2.0$.

Finally, for verification of the wind speed forecasts, we employed the fair ensemble CRPS \citep{Leutbecher}, which allows for a distribution-free evaluation of ensemble predictions:
\begin{equation}
\text{CRPS}_{\text{fair}} = \frac{1}{m} \sum_{i=1}^{m} |x_i - y| - \frac{1}{2 m (m - 1)} \sum_{i=1}^{m} \sum_{\substack{j=1 \\ j \ne i}}^{m} |x_i - x_j|.
\end{equation}

\section{Supplementary results} \label{appendix:C} 
\subsection{Spread-error ratio} 
\begin{figure}[H]
	\centering
	\includegraphics[scale=0.35]{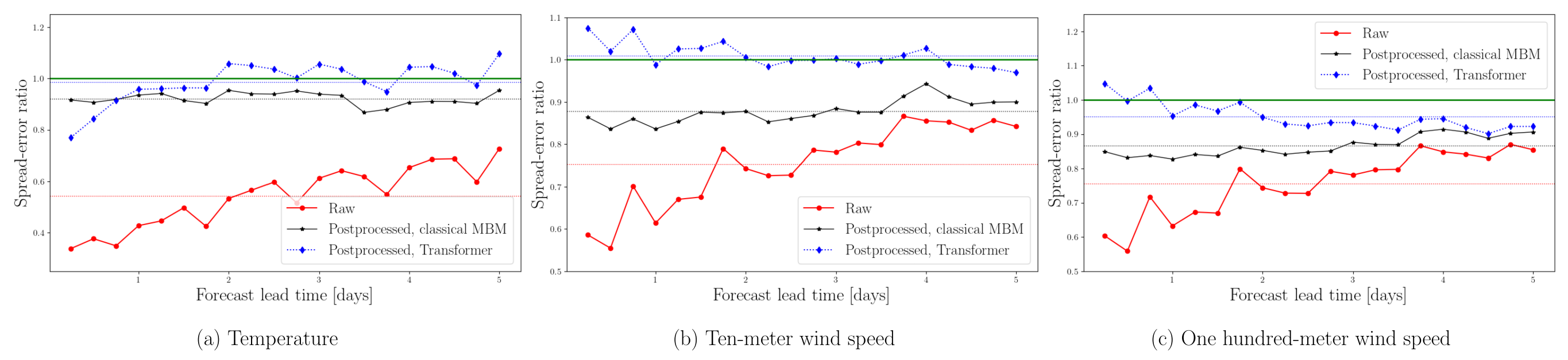}
	\caption{Spread-error ratio (SER) per lead time for the postprocessed forecasts of (a) two-meter temperature, (b) ten-meter wind speed and (c) one hundred-meter wind speed. A SER value closer to one is better, because that means the ensemble spread and the root-mean-squared error are in balance.}
	\label{fig:reslead}
\end{figure}

Figure \ref{fig:reslead} presents the average SER values per lead time.

\subsection{Bias} 
\begin{figure}[H]
	\centering
	\includegraphics[scale=0.35]{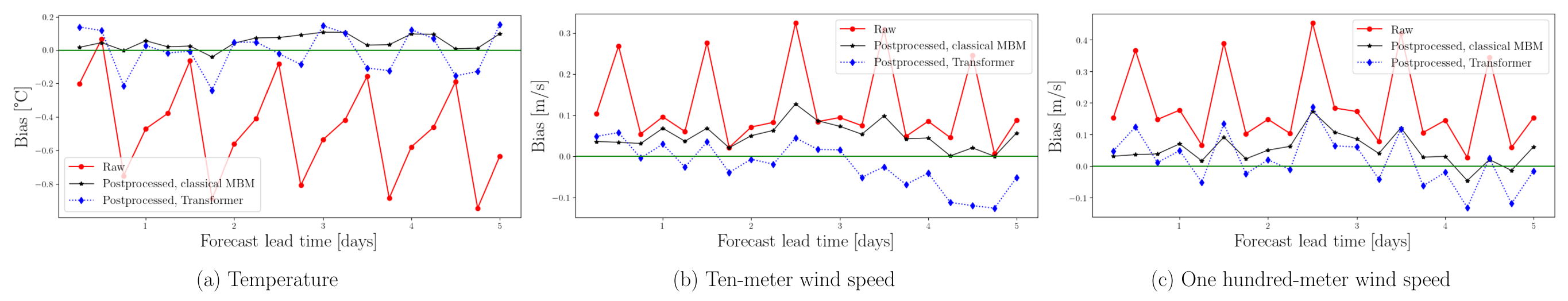}
	\caption{Bias per lead time for the postprocessed forecasts of (a) two-meter temperature, (b) ten-meter wind speed and (c) one hundred-meter wind speed.}
	\label{fig:biaslead}
\end{figure}

\begin{figure}[H]
	\centering
	\includegraphics[scale=0.65]{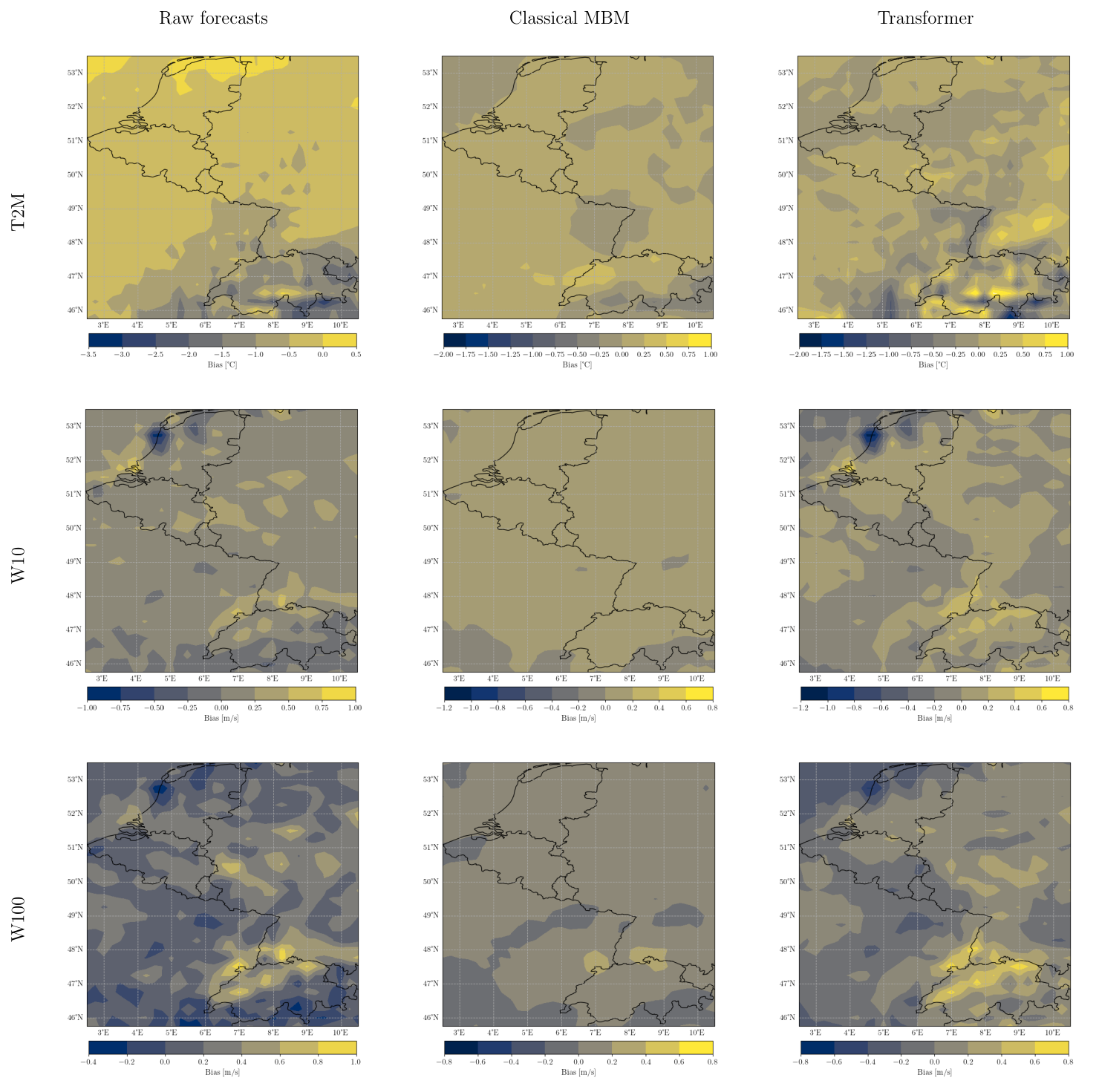}
	\caption{Bias for two-meter temperature (\textit{top row}), ten-meter wind speed (\textit{middle row}) and one hundred-meter wind speed (\textit{bottom row}) over the spatial domain, averaged over all lead times, for the raw forecasts (\textit{left column}) and forecasts postprocessed by classical MBM (\textit{middle column}) and by the Transformer (\textit{right column}). For clarity, the scale of the raw forecasts differs from that of the images for classical MBM and the Transformer.}
	\label{fig:biasgrid}
\end{figure}

Figure \ref{fig:biaslead} presents the average bias per lead time, while Figure \ref{fig:biasgrid} represents the bias of every method over the EUPPBenchmark dataset, averaged over the lead times. 
\subsection{Ensemble example} 
\begin{figure}[h!]
	\centering
	\includegraphics[scale=0.35]{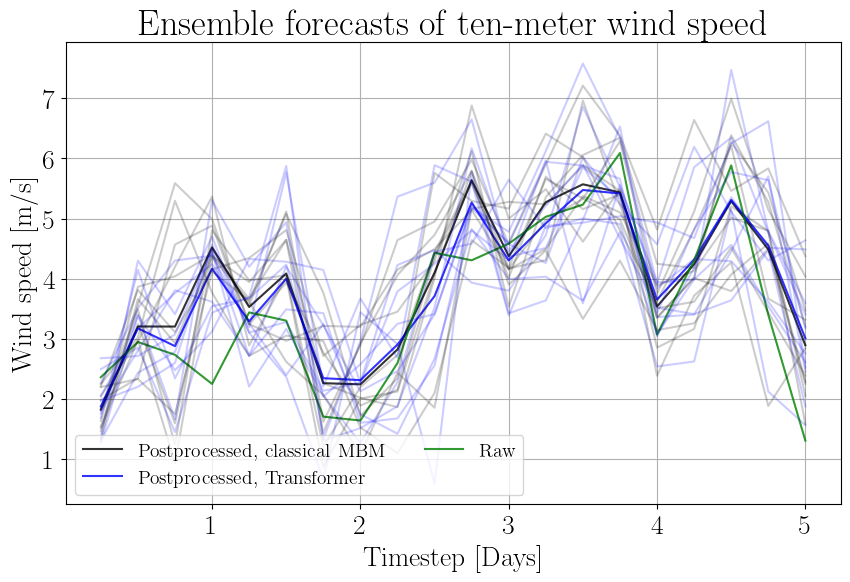}
	\caption{An ensemble of ten-meter wind speed forecasts, for a random day and a random grid point in function of lead time.}
	\label{fig:EnsembleExample}
\end{figure}

An example of an ensemble of ten-meter wind speed forecasts as postprocessed by the Transformer and classical MBM is presented in Figure \ref{fig:EnsembleExample}.

\bibliographystyle{ametsocV6}
\bibliography{Postprocessing_Transformers_Arxiv_update}

\end{document}